\documentclass{llncs}
\usepackage{times}

\usepackage{soul}
\usepackage{url}
\usepackage{hyperref}
\hypersetup{hidelinks}
\usepackage[utf8]{inputenc}
\usepackage[small]{caption}
\usepackage{graphicx}
\usepackage{amsmath}
\usepackage{booktabs}
\usepackage{algorithm}
\usepackage{todonotes}
\usepackage{tikz}
\usepackage{amssymb}
\usetikzlibrary{shapes,arrows,chains}
\usetikzlibrary[calc]
\usepackage{comment}

\usepackage{placeins}
\definecolor{dartmouthgreen}{rgb}{0.05, 0.5, 0.06}

\newtheorem{principle}{Principle}

\newenvironment{exs}[1]{\begin{trivlist}\item[]\textit{Example \ref{#1} (Continued).}}{\end{trivlist}} 

\newcommand{\cP}{{\mathcal{P}}}

\newcommand{\cS}{{\mathcal{S}}}

\newcommand{\bC}{{\mathbf{C}}}

\newcommand{\bF}{{\mathbf{F}}}

\newcommand{\bI}{{\mathbf{I}}}

\newcommand{\bM}{{\mathbf{M}}} 
 
 \newcommand{\bfo}{{\mathbf{o}}}
\newcommand{\bP}{{\mathbf{P}}} 
 
\newcommand{\bR}{{\mathbf{R}}} 
\newcommand{\bS}{{\mathbf{S}}} 
 \newcommand{\bft}{{\mathbf{t}}}
 
\newcommand{\bV}{{\mathbf{V}}}

\newcommand{\card}[1]{|#1|}
\newcommand{\tuple}[1]{\langle #1 \rangle}

\newcommand{\CN}{{\mathtt{Cn}}}

\newcommand{\conp}{\mathtt{pCon}}
\newcommand{\conf}{\mathtt{tCon}}
\newcommand{\incp}{\mathtt{pInc}}
\newcommand{\incf}{\mathtt{tInc}}

\newcommand{\sortfunc}{\mathtt{sort}}
\newcommand{\arity}{\mathtt{ar}}

\newcommand{\sort}{\bS\bfo}
\newcommand{\logic}{\mathtt{MS}$-$\mathtt{FOL}}
\newcommand{\struct}{\bS\bft}

\newcommand{\MI}{\mathtt{MI}}
\newcommand{\TMLN}{\mathtt{TMLN}}
\newcommand{\TI}{\mathtt{TI}}
\newcommand{\TF}{\mathtt{TF}}
\newcommand{\W}{\mathtt{W}}

\newcommand{\tlogic}{\mathtt{TS}$-$\mathtt{FOL}}

\newcommand{\ie}{{\textit{i.e., }}}
\newcommand{\wrt}{{\textit{w.r.t. }}}

\newcommand{\bigsig}{\scalebox{1.5}{$\sigma$}}
\newcommand{\bigtau}{\scalebox{1.2}{$\tau$}}

\title{Parameterisation of Reasoning on Temporal Markov Logic Networks}

\author{
Victor David$^1$\and
Raphaël Fournier-S'niehotta $^1$\and
Nicolas Travers$^{1,2}$}
\institute{
$^1$Conservatoire National des Arts et Métiers\\
$^2$Léonard de Vinci Pôle Universitaire, Research Center}

\begin{document}
	
%
\maketitle  

%
%

\begin{abstract}
We aim at improving reasoning on inconsistent and uncertain data. We focus on knowledge-graph data, extended with time intervals to specify their validity, as regularly found in historical sciences.
We propose principles on semantics for efficient Maximum A-Posteriori inference on the new Temporal Markov Logic Networks (TMLN) which extend the Markov Logic Networks (MLN) by uncertain temporal facts and rules.
We examine total and partial temporal (in)con\-sistency relations between sets of temporal formulae. 
Then we propose a new Temporal Parametric Semantics, which may combine several sub-functions, allowing to use different assessment strategies.
Finally, we expose the constraints that semantics must respect to satisfy our principles. 




%
%
\end{abstract}



\section{Introduction}

Reasoning on large data sets to extend knowledge is an open challenge~\cite{10.5555/1625275.1625705,10.1145/1409360.1409378,10.1007/978-3-642-41335-3_34,10.1145/3442442.3458603}.
Most approaches model information with \textit{Knowledge Graphs} (KGs)~\cite{10.1145/2623330.2623623}, and rely on \textit{Ontologies}~\cite{10.1145/3091478.3162385}, 
\textit{Machine Learning}~\cite{10.1145/3447548.3470825} or \textit{Neural Networks}~\cite{10.1145/3340531.3412164} representations.
Then, \textit{Description Logic}~\cite{10.1145/3133811.3133830} and \textit{Temporal Logic}~\cite{10.1145/2883817.2883839} are used for rules satisfiability.

For historians, reasoning under uncertainty and inconsistency is at the basis of methodology: the validity of any fact remains questionable. Temporal information is crucial: outside of a temporal interval, a fact becomes false.
Using facts, temporal and uncertainty information, a historian may obtain several consistent levels of knowledge.

Recent works on KG reasoning do not focus on both time and uncertainty. 
Several models with \textit{Markov Logic Networks} (MLN) have been proposed~\cite{10.3233/978-1-61499-672-9-1017,10.1007/978-3-319-99906-7_10}. One focused on reasoning on \textit{Uncertain Temporal Knowledge Graphs} (UTKG) with specific temporal inference rules~\cite{chekol2017marrying}.
%
However, those representations cannot handle a fully uncertain universe where any fact or rule may be uncertain.

We propose to generalize this approach, by 
i) introducing a new representation of knowledge, called \emph{temporal MLN} (TMLN), built on (a temporal) many-sorted logic, for reasoning on both time and uncertainty, 
ii) enlarging the notion of uncertainty to 
rules, and
iii) adapting reasoning, to deal with both uncertain facts and rules. We define a new temporal semantics and a temporal extension to \textit{Maximum A-Posteriori} (MAP) inference~ \cite{riedel2012improving}.
This MAP inference produces \emph{instantiations}, \ie\,extended sets of facts maximizing the score \wrt a temporal semantics.
Finally, we generalise the temporal semantics by a Parametric Semantics, which may combine several sub-functions for various consistency validations.


\section{Background}



In a seminal work~\cite{chekol2017marrying}, Chekol et al. formalise the UTKG approach, which integrates both time and uncertainty in Knowledge Graphs to reach a \textit{certain world maximisation}.
However, they do not take into account the possibility to have uncertain rules.
We put \emph{time} at the heart of reasoning, and enlarge their vision. 
Firstly, we formalise the notion of temporal uncertainty, by combining certain and uncertain 
formulae.
Secondly, our novel representation of TMLN allows for easier manipulations and better analyses.

\subsection{Many-Sorted First Order Logic}
\label{sec:msfol}

We start by presenting what is a Many-Sorted First-Order Logic (introduced in  \cite{wang1952logic}). 

\begin{definition}[Many-Sorted FOL]
Let $\sort = \{s_1,\dots,s_n\}$ be a set of sorts.
A {\em Many-Sorted First-Order Logic} $\logic$, is a set of formulae built up by induction from:
    a set $\bC = \{a_1, \ldots, a_l\}$ of constants, 
    a set $\bV = \{x^s,y^s,z^s, \ldots \mid s \in \sort\}$ of variables,
    a set $\bP = \{P_1, \ldots, P_m\}$ of predicates,
    a function $\arity: \bP \rightarrow \mathbb{N}$ which tells the arity of any predicate,
    a function $\sortfunc$ s.t. for $P \in \bP$, $\sortfunc(P) \in \sort^{\arity(P)}$, and for $c \in \bC$, $\sortfunc(c) \in \sort$,
    the usual connectives ($\neg$, $\vee$, $\wedge$, $\rightarrow$, $\leftrightarrow$), Boolean constants ($\top$ and $\bot$) and quantifier symbols ($\forall, \exists$).
A {\em ground formula} is a formula without any variable.
\end{definition}

Lowercase (resp. uppercase) greek letters like $\phi, \psi$ (resp. $\Phi, \Psi$) denote formulae (resp. sets of formulae). $\logic$ is the set of all formulae.
%

\begin{example}
For instance let $\sort = \{s_1,s_2\}$, let $P_1 \in \bP$ such that $\sortfunc(P_1) = s_2 \times s_1 \times s_1$, let $a_1,a_2,t_1,t_2 \in \bC$ such that $\sortfunc(a_1) = \sortfunc(a_2) = s_2$, $\sortfunc(t_1) = \sortfunc(t_2) = s_1$ and let $x^{s_2} \in \bV$. 
We can build then the following $\logic$ formulae: $P_1(a_1,t_1,t_2)$, $\forall x^{s_2} P_1(x^{s_2},t_1,t_2)$. However, $P_1(t_1,t_2,a_1)$ or $\forall x^{s_2} P_1(a_1,a_2,x^{s_2})$ cannot be built because they do not respect the sorts.
\end{example}

$\logic$ formulae are evaluated via a notion of \textit{structure} called $n$-sorted structures~\cite{gallier2015logic}.
Classical first-order logic formulae are captured as $1$-sorted structures.

    
\begin{definition}[Structure]
A {\em $n$-sorted structure} is $\struct= (\{D_1,\ldots,D_n\}, \{R_1,\ldots,R_m\}, \linebreak \{c_1,\ldots,c _l\})$, where $D_1,\ldots,D_n$ are the (non-empty) domains, $R_1,\ldots,R_m$ are relations between domains' elements, and $c_1,\ldots,c_l$ are distinct constants in the domains.
\end{definition}



Our running example is presented in Example~\ref{ex:running}. These sentences gather biographical elements about a French philosopher from the 14th century, \textit{Nicole Oresme}.

\begin{example}\label{ex:running}
Nicole Oresme was a person and a philosopher born in the Middle Ages between 1320-1382.
It would appear that Nicole Oresme attended the College of Navarre around 1340-1354 and more likely around 1355-1360.
Nicole Oresme possibly did not attend the College of Navarre around 1353-1370.
Sometimes, a person who lived in Middle Ages and studied at College of Navarre came from a peasant family.
Usually, a philosopher born in the Middle Ages did not come from a peasant family.
\end{example}

In $\logic$, though without uncertainty, we may then define a suitable structure.

\begin{example} \label{ex:struct}
An example of structure associated with the $\logic$ from Example~\ref{ex:running} 
is $\struct_{hist}$\,=\,$(\{Time, Concept\},$ $\{Person, Philosopher, LivePeriod, PeasantFamily, \linebreak Studied\},$ $\{ t_{min}, 1300, 1301, 1302, \ldots , 1400, t_{max}, NO, MA, CoN\})$, in which:

\begin{itemize}
    \item 
    $Time$ is the set of time points, corresponding to the sort $s_1$ and $Concept$ is the set of all non-temporal objects, corresponding to the sort $s_2$, 
    
    \item 
    $Person, Philosopher, LivePeriod$, etc. are the predicate symbols' relations  ({\em e.g.,} $Person \subseteq Concept \times Time \times Time$ indicates which elements are a person). 
    
    \item  
    $t_{min}$, 1300, 1301, $\ldots$, 1400, $t_{max}$ are elements of the domain $Time$ associated with the sort $s_1$, while $NO$ (Nicolas Oresme), $MA$ (Middle Ages) and $CoN$ (College of Navarre) are elements of the domain $Concept$ associated with the sort $s_2$.
\end{itemize}
\end{example}


Now, we show how $\logic$ formulae are \emph{interpreted}.

\begin{definition}[Interpretation]\label{def:ms-fol-interpretation}
An {\em interpretation $\bI_{\struct}$ over a structure $\struct$} assigns to elements of the $\logic$ vocabulary some values in the structure $\struct$. Formally,\\
    -- $\bI_{\struct}(s_i) = D_i$, for $i \in \{1,\dots,n\}$ 
    (each sort symbol is assigned to a domain), \\ 
    -- $\bI_{\struct}(P_i) = R_i$, for $i \in \{1,\dots,m\}$ (each predicate symbol is assigned to a relation),\\
    -- $\bI_{\struct}(a_i) = c_i$, for $i \in \{1,\dots,l\}$ (each constant symbol is assigned to a value).\\
Then, satisfying formulae is recursively defined by:\\
%
    -- $\bI_{\struct} \models P_i(a_1,\dots,a_k)$ iff $(\bI_{\struct}(a_1),\dots,\bI_{\struct}(a_k)) \in R_i$, \\
    -- $\bI_{\struct} \models \exists x^{s_i} \phi$ iff $\bI_{\struct, x^{s_i} \leftarrow v} \models \phi$ for some $v \in D_i$, \\
    -- $\bI_{\struct} \models \forall x^{s_i} \phi$ iff $\bI_{\struct, x^{s_i} \leftarrow v} \models \phi$ for each $v \in D_i$,\\
    -- $\bI_{\struct} \models \phi \wedge \psi$ iff $\bI_{\struct} \models \phi$ and $\bI_{\struct} \models \psi$,\\
    -- $\bI_{\struct} \models \phi \vee \psi$ iff $\bI_{\struct} \models \phi$ or $\bI_{\struct} \models \psi$,\\
    -- $\bI_{\struct} \models \neg \phi$ iff $\bI_{\struct} \not\models \phi$.\\
where $\bI_{\struct, x^{s_i} \leftarrow v}$ is a modified version of $\bI_{\struct}$ s.t. the variable $x^{s_i}$ is replaced by a value $v$ in the domain $D_i$ corresponding to the sort symbol $s_i$.
Finally, if $\Phi$ is a set of formulae, then $\bI_{\struct} \models \Phi$ iff $\bI_{\struct} \models \phi$ for each $\phi \in \Phi$.
\end{definition}

Definition~\ref{def:ms-fol-interpretation} does not target the satisfaction of implications and equivalences, while they can be defined by: $(\phi \rightarrow \psi) \equiv (\neg \phi \vee \psi)$, and $(\phi \leftrightarrow \psi) \equiv (\phi \rightarrow \psi) \wedge (\psi \rightarrow \phi)$.
For instance, the set of interpretations of the formula $P(a) \vee P(b)$ is equal to $\{\{P(a)\}, \{P(b)\}, \{P(a),P(b)\}\}$ and for $P(a) \wedge P(b)$ is $\{\{P(a),P(b)\}\}$.\\


With these notions of structures and interpretations on TMLNs, we may define the consequence relations and logical consequences over $\logic$ formulae.

\begin{definition} [Consequence Relation] \label{def:ms-fol-consequence-relation}
Let $\phi$ and $\psi$ be two $\logic$ formulae. We say that {\em $\psi$ is a consequence of $\phi$}, denoted by $\phi \vdash \psi$, if for any structure $\struct$, and any interpretation $\bI_{\struct}$ over $\struct$, $\bI_{\struct} \models \phi$ implies $\bI_{\struct} \models \psi$. 
\end{definition}

\begin{definition} [Logical Consequences - $\CN$]\label{def:fol_lc}
Let $\phi \in \logic$.
The function $\CN(\phi)$ is the set of all logical consequences of $\phi$, i.e. $\CN(\phi) = \{\psi \in \logic \ | \ \phi \vdash \psi\}.$
\end{definition}
The function $\CN$ returns an infinite set of formulae, but for clarity we consider only one formula per equivalent class and only the predicates and constants appearing in the original formulae. 
Such as, $\CN(P(a) \vee P(b)) = \{P(a) \vee P(b)\}$ and\\ $\CN(P(a) \wedge P(b)) = \{P(a), P(b), P(a) \wedge P(b), P(a) \vee P(b)\}\}$.

\section{Temporal and Uncertain Knowledge Representation}

Markov Logic Networks (MLNs) combine Markov Networks and First-Order Logic (FOL) by attaching weights to ﬁrst-order formulae and treating them as feature templates for Markov Networks~\cite{Richardson2006}.
We extend this framework to temporal information by resorting to Many-Sorted  First-Order Logic ($\logic$).

\subsection{Temporal Markov Logic Networks}

Let start by defining what we call a Temporal Many-Sorted First-Order Logic $\tlogic$.
\begin{definition}[Temporal Many-Sorted FOL] 
A $\tlogic$ evaluated by a structure $\struct$ is a constrained  $\logic$ where $\card{\sort} \geq 2$, for any interpretation $\bI_{\struct}(s_1) = Time$, any predicate $P_i \in \tlogic$ has $\arity(P_i) \geq 3$ with the sort of the last two parameters belonging to $s_1$ and $t_{min}$ and $t_{max}$ are  time constants indicating the minimum and maximum time points for any pre-order between the time constants. 


\end{definition}

Using this constrained $\logic$ accompanied with a temporal domain (named $Time$) and temporal predicates (the last two parameters indicate the validity temporal bounds), we may represent temporal facts and rules.
Finally, Temporal Markov Logic Networks (TMLN) extend $\tlogic$ (resp. Markov Logic Networks) by associating a degree of certainty to each formula (resp. by adding a temporal validity to the predicates).
 

\begin{definition}[TMLN] \label{def:tmln}
\noindent A Temporal Markov Logic Network $\bM = (\bF,\bR)$, based on a $\tlogic$, is a set of weighted temporal facts and rules where $\bF$ and $\bR$ are sets of pairs such that:\\
-- $\bF = \{(\phi_1, w_1), $\ldots$, (\phi_n, w_n)\}$ with $\forall i \in \{1, $\ldots$, n\}$, $\phi_i \in \tlogic$ such that it is a ground formula and $w_i \in [0,1]$,\\
-- $\bR = \{(\phi'_1, w'_1), $\ldots$, (\phi'_k, w'_k)\}$ with $\forall i \in \{1, $\ldots$, k\}$, $\phi'_i \in \tlogic$ such that it is not a ground formula and in the form (premises,conclusion), i.e., $(\psi_1 \wedge $\ldots$ \wedge  \psi_l) \rightarrow \psi_{l+1}$ where $\forall j \in \{1,$\ldots$, l+1\}$, $\psi_j \in \tlogic$, and $w_i \in [0,1]$.\\
The universe of all TMLNs is denoted by $\mathtt{TMLN}$.
\end{definition}





\begin{table*}[t]
    \centering
    \scriptsize
    \begin{tabular}{clll}
        $F_1$ & $(Person(NO, 1320,1382)$&&$, \textbf{1})$ \\
        $F_2$ & $(Philosopher(NO, 1320,1382)$&&$, \textbf{1})$\\
        $F_3$ &$(LivePeriod(NO, MA,  1320,1382)$&&$, \textbf{1})$\\
        $F_4$&$(Studied(NO, CoN, 1340,1354)$&&$, \textbf{0.4})$\\
        $F_5$&\multicolumn{2}{l}{$(Studied(NO, CoN, 1355,1360)$}&$, \textbf{0.7})$\\
        $F_6$ & \multicolumn{2}{l}{$(\neg Studied(NO, CoN, 1353,1370)$}&$, \textbf{0.5})$\\
        $R_1$&\multicolumn{2}{l}{$(\forall x^{s_2},t_1^{s_1},t_1'^{s_1},t_2^{s_1},t_2'^{s_1},t_3^{s_1},t_3'^{s_1}, (Person(x^{s_2}, t_1^{s_1},t_1'^{s_1}) \wedge LivePeriod(x^{s_2},MA, t_2^{s_1},t_2'^{s_1}) \wedge$} & \\
        &\hspace*{1cm} $Studied(x^{s_2}, CoN, t_3^{s_1},t_3'^{s_1})) \rightarrow PeasantFamily(x^{s_2},t_{min},t_{max})$ &&$, \textbf{0.5})$\\
        $R_2$& $(\forall x^{s_2},t_1^{s_1},t_1'^{s_1},t_2^{s_1},t_2'^{s_1}, (Philosopher(x^{s_2}, t_1^{s_1},t_1'^{s_1}) \wedge LivePeriod(x^{s_2}, MA, t_2^{s_1},t_2'^{s_1}))$&\\
        &\hspace*{4.7cm}$\rightarrow \neg PeasantFamily(x^{s_2},t_{min},t_{max})$&&$, \textbf{0.8})$\\
    \end{tabular}
    \caption{Example of a TMLN for Nicole Oresme.}
    \label{tab:TMLN}
\end{table*}

\begin{table*}[t]
    \centering
    \scriptsize
    \begin{tabular}{ll}
$GR_{11} = ((Person(NO, 1320,1382) \wedge LivePeriod(NO,MA, 1320,1382) \wedge$\\
\hspace*{1cm}$Studied(NO, CoN, 1340,1354))
 $
 \hspace*{1.5cm}
 $\rightarrow PeasantFamily(NO,t_{min},t_{max})$& , \textbf{0.4})\\
$GR_{12} = ((Person(NO, 1320,1382) \wedge LivePeriod(NO,MA, 1320,1382) \wedge$\\
\hspace*{1cm}$Studied(NO, CoN, 1355,1360))
 $
 \hspace*{1.5cm}
 $\rightarrow PeasantFamily(NO,t_{min},t_{max})$& , \textbf{0.5})\\
$GR_2 = ( (Philosopher(NO, 1320,1382) \wedge LivePeriod(NO, MA, 1320,1382))$\\
\hspace*{6.5cm}$ \rightarrow \neg PeasantFamily(NO,t_{min},t_{max})$& , \textbf{0.8})\\
\end{tabular}
    \caption{Ground Rules Instantiating $R_1$ and $R_2$ (from Table~\ref{tab:TMLN}) for Nicole Oresme.}
    \label{tab:GR}
\end{table*}

In the rest of the paper, we simplify the example by directly using the structure defined in Example~\ref{ex:struct} (\textit{c.f.} Section~\ref{sec:msfol}). 

\begin{exs}{ex:running}
The TMLN representation of our running example can be found in 
Table~\ref{tab:TMLN}. 
We may identify 6 independent facts and 2 rules, each one with temporal validity and certainty weights (arbitrary extracted from Example~\ref{ex:running}).
\end{exs}

In order to select what is the most probable and consistent set of ground formulae with a MAP inference, we need first to have our data (facts and rules) represented in a TMLN. 
Then, we must obtain the ground rules (if possible), by replacing the variables in the rules by constants (according to our TMLN). We call this second step the instantiation.


\subsection{TMLN Instantiation} \label{sec:instantiation}
Let $\bM$ be a TMLN, 
we denote by $\MI(\bM)$ the \textit{Maximal TMLN Instantiation} of $\bM$. 
$\MI(\bM)$ contains the set of $\bM$'s facts and all ground rules that can be constructed by instantiating all its predicates containing variables by other deductible ground predicates (with logical consequence, from  Definitions~\ref{def:ms-fol-consequence-relation} and~\ref{def:fol_lc}).  
A ground rule's weight is the minimum of the weights of the formulae in $\bM$ used to construct the instantiated rule.
%

Formally, to define the set of instantiations, we have to define two useful notions. Firstly, we denote by $\TF(\bM) = \bigcup\limits_{(\phi,w) \in \bM} \phi$ the set of temporal formulae of $\bM \in \mathtt{TMLN}$. 
%
 Secondly, we define the function $\W : \tlogic \times \TMLN \rightarrow [0,1]$,  returning the maximal weight of a temporal formula deductible from a TMLN:
 $\mathtt{W}(\phi, \bM ) = \mathtt{max}(\mathtt{min_w}(\bM_1), \ldots, \linebreak  \mathtt{min_w}(\bM_m))$ s.t. $\{ \bM_1, \ldots, \bM_m \} = \{ \bM_i \subseteq \bM \mid \TF(\bM_i) \vdash \phi \text{ and } \nexists \bM'_i \subset \bM_i \text{ s.t. } \linebreak  \TF(\bM'_i) \vdash \phi \}$ and $\mathtt{min_w}(\bM_i = \{(\psi_1,w_1),\ldots,(\psi_l,w_l)\} ) = \mathtt{min}(w_1, \ldots, w_l)$.

\begin{definition}[TMLN Instantiation]\label{definition:instantiation}
Given $\bM = (\bF,\bR) \in \TMLN$, 
the set of {\em instantiations} $\MI$ of $\bM$ is defined as follows:

$\MI(\bM) = \bF ~~ \cup \{\big((\phi_1 \wedge \ldots \wedge \phi_k \rightarrow \phi_{k+1})_{X \leftarrow C},w'\big) \mid (\phi_1 \wedge \ldots \wedge \phi_k \rightarrow \phi_{k+1},w) \in \bR, X = \{x_1, \ldots, x_n\} \text{ is the set of }   \text{variables in } \phi_1 \wedge \ldots \wedge \phi_k \rightarrow \phi_{k+1}, C = \tuple{c_1,\ldots, c_n} \text{ is a vector of constants replacing each occurrence of the variables, } \\ X'_i \subseteq X \text{ is the set of variables in } \phi_i, C'_i \subseteq C \text{ is the vector of constants replaced in } \phi_i \\ \text{ and the instantiated rule satisfies the 2 following conditions:}$

        1. $\forall \phi_i \in \{\phi_1, \ldots, \phi_{k}\}$, $\phi_{i X'_i \leftarrow C'_i} \in \CN(\TF(\bM))$
        
        2. $w' =  \mathtt{min}(w, \mathtt{W}(\phi_{1 X'_1 \leftarrow C'_1}, \bM),\ldots, \mathtt{W}(\phi_{k X'_k \leftarrow C'_k}, \bM ) \}$\\
Where $\phi_{X \leftarrow C}$ is the formula $\phi$ s.t. all the occurrences of the variable $x_i \in X$ are replaced by the constant $c_i \in C$.

\end{definition}

Currently, we only deal with universal (\textit{i.e.,} $\forall$) rules and no existential one (\textit{i.e.,} $\exists$), to simplify the maximal TMLN instantiation. Indeed, with existential rules, we would have to deal with a set of sets of instantiations, given that we would not know which set of instantiations would be true. We keep this question for future works.

From Example~\ref{ex:running}, the instantiation of $R_1$ (resp. $R_2$) are $GR_{11}$ (from $F_1$, $F_3$ and $F_4$)  and $GR_{12}$ (from $F_1$, $F_3$ and $F_5$) (resp. $GR_2$ from $F_2$ and $F_3$). Hence $GR_{11}$ has a weight of $0.4$, $GR_{12}$ of $0.5$ and $GR_{2}$ of $0.8$, 
see Table~\ref{tab:GR}. 

A TMLN instantiation $I \subseteq \MI(\bM)$ is a TMLN only composed of ground formulae, $I$ is also called \textit{a state of the TMLN $\bM$}.  The universe of all TMLN instantiations is denoted by $\TMLN^*$. 
Note that an instantiation can be inconsistent. In our example, $GR_{11}, F_1, F_3, F_4$ imply $PeasantFamily(NO,t_{min}, t_{max})$ while $GR_2, F_2, F_3$ imply $\neg PeasantFamily(NO,t_{min}, t_{max})$, \textit{i.e.}, they are inconsistent together.
Thus, to obtain the most consistent and informative set of instantiations, we have to compute the \textit{Maximum A-Posteriori} (MAP) inference~\cite{chekol2017marrying,riedel2012improving}.

\section{Temporal and Uncertain Knowledge Reasoning}

Computing the MAP inference means ``finding the most probable state of the world''~\cite{chekol2017marrying}. In order to do that, we integrate semantics on TMLN, before examining principles governing those semantics and (in)consistency relations.

\subsection{Temporal MAP Inference}
\label{sec:map}
%
 \textbf{Semantics} 
are methods 
which compute the strength of a TMLN state. We denote the universe of all semantics by $\mathtt{Sem}$, such that for any $\cS \in \mathtt{Sem}$,  
$\cS : \mathtt{TMLN^*} \rightarrow [0, +\infty[.$ 
We compute a strength above 0, instead of a probability between 0 and 1.
One semantics may maximise the amount of information, while another may maximise the quality. 


\noindent \textbf{Temporal Maximum A-Posteriori (MAP) Inference} 
in TMLN returns the most probable, temporally consistent, and expanded state \textit{w.r.t.} a given semantics.
Given $\bM \in \mathtt{TMLN}$ and $\cS \in \mathtt{Sem}$, 
a method solving a MAP problem is denoted by:\\ $\mathtt{map} : \mathtt{TMLN} \times \mathtt{Sem} \rightarrow \cP(\mathtt{TMLN^*})$, where $\cP(X)$ denote the powerset of X, such that:

\noindent $\mathtt{map}(\bM,\cS) = \{ I \mid I \in \underset{I ~\subseteq ~\MI(\bM) }{\text{argmax}}~\cS(I) \text{ and } \nexists I' \in \underset{I' ~\subseteq ~\MI(\bM) }{\text{argmax}}~\cS(I') \text{ s.t. } I \subset I'\}$.


To determine a MAP inference we need to define semantics, however not all methods are desirable. Below we present some principles that a semantics should satisfy.

\subsection{Principles for semantics in Temporal MAP Inference} 
\label{sec:principles}







Our first principle states that adding new weightless information does not change the strength of the MAPs of a TMLN, whatever the temporality is.
Given that the same set of predicates or formulae instantiated on different temporalities are not equivalent, we homogenises $\tlogic$ with the maximal time interval to define the information novelty.

For time homogenization, we denote by $\bigtau : \cP(\tlogic) \rightarrow \cP(\tlogic)$, the function transforming any temporal predicate into the maximal time interval ($t_{min}$ and $t_{max}$).  

\begin{principle}[Temporal Neutrality] \label{prin:TN}
Let $\bM \in \mathtt{TMLN}$, $\bM' = \bM \cup \{(\phi,w)\}$ where $(\phi,w)$
is a weighted temporal formula ($\phi \in \tlogic$ and $w \in [0,1]$), such that:

    -- $\bigtau(\TF(\bM)) \not\vdash \bigtau(\{\phi\})$, and
    
    -- $w = 0$.

\noindent A semantics $\cS \in \mathtt{Sem}$ satisfies temporal neutrality iff, 
$\forall I \in \mathtt{map}(\bM,\cS)$ and $\forall I' \in \mathtt{map}(\bM',\cS)$, $\cS(I) = \cS(I')$.
\end{principle}

The next principle ensures that one cannot decrease the strength of the MAPs of a TMLN by adding new and consistent information. 
%
The temporal consistency $\mathtt{Con}: \cP(\tlogic) \rightarrow \{\top,\bot\}$ denotes a relation of consistence for a set of temporal formulae.

\begin{principle}[Consistency Monotony]  \label{prin:mon}
Let a relation of consistence $\mathtt{Con}$, $\bM \in \mathtt{TMLN}$ and  $\bM' = \bM \cup \{(\phi,w)\}$ where $(\phi,w)$ is a weighted temporal formula such that: 

    -- $\bigtau(\TF(\bM)) \not\vdash \bigtau(\{\phi\})$, and
    
    -- $\forall I \in \mathtt{map}(\bM,\cS)$, $\mathtt{Con}(\{\phi\} \cup \TF(I))$ is true and $I \subset \MI(\{(\phi,w)\} \cup I)$.\\
A semantics $\cS \in \mathtt{Sem}$ satisfies consistency monotony iff: 
$\forall I \in \mathtt{map}(\bM,\cS)$ and \linebreak $\forall I' \in \mathtt{map}(\bM',\cS)$, 
$\cS(I) \leq \cS(I')$. 
\end{principle}




The last principle states that if we add a new temporal fact to a TMLN such that it is consistent with each instantiation, then the fact will be present in each new instantiation.

\begin{principle}[Invariant Consistent Facts]  \label{prin:fact}
Let a relation of consistence $\mathtt{Con}$, $\bM \in \mathtt{TMLN}$ and  $\bM' = \bM \cup \{(\phi,w)\}$ where $(\phi,w)$ is a TMLN fact  such that:

    -- $\bigtau(\TF(\bM)) \not\vdash \bigtau(\{\phi\})$, and
    
    -- $\forall I \in \mathtt{map}(\bM,\cS)$, $\mathtt{Con}(\{\phi\} \cup \TF(I))$ is true.\\
A semantics $\cS \in \mathtt{Sem}$ satisfies invariant consistent facts iff, 
$\forall I \in \mathtt{map}(\bM,\cS)$,  $I \cup \{(\phi,w)\} \in \mathtt{map}(\bM',\cS)$. 
\end{principle}

\subsection{Temporal Consistency and Inconsistency}
\label{sec:TC_TIc}
We study here new temporal consistency interactions required to define our \textit{Temporal MAP inference}.
%
\textit{Temporal Consistency} relations need to be refined according to the temporal validity of the predicates.
For a predicate and its negation, no clear definition exists to express the temporal consistencies based on their time intervals.
We propose a temporal consistency with a general case (\textit{partial}) and a special case (\textit{total}).

To establish these different temporal consistency relations, we introduce a function \texttt{TI} to create pre-orders between the temporal constants in the domain $Time$ of a $\tlogic$ and which extracts the interval of time points from two constants.


\begin{definition}[Temporal (in)consistency] \label{def:ticon}
Let a set of formulae $\Phi \subseteq \tlogic$.

\noindent \textbf{Temporal consistency:}

        -- $\Phi$ has a partial temporal consistency 
denoted by $\conp(\Phi)$ iff: 
$\forall \phi, \psi \in \CN(\Phi) \text{ s.t. }$

$~~~ \phi = P(x_1,$\ldots$,x_k,t_1,t_1') \text{ and } \psi = \neg P(x_1,$\ldots$,x_k,t_2,t_2') \text{, } (\TI(t_1,t_1') \setminus \TI(t_2,t_2')$

$~~\neq \emptyset) \wedge (\TI(t_2,t_2') \setminus \TI(t_1,t_1') \neq \emptyset). $

$~~$ Otherwise $\neg \conp(\Phi)$ is true. 

        -- $\Phi$ has a total temporal consistency 
denoted by $\conf(\Phi)$ iff: 
\noindent$\forall \phi, \psi \in \CN(\Phi) \text{ s.t. } \phi =$

$~~~ P(x_1,$\ldots$,x_k,t_1,t_1') \text{ and } \psi = \neg P(x_1,$\ldots$,x_k,t_2,t_2') \text{, } (\TI(t_1,t_1') \cap \TI(t_2,t_2')$=$\emptyset).$

$~~$ Otherwise $\neg \conf(\Phi)$ is true.  


\noindent \textbf{Temporal inconsistency:}

        -- $\Phi$ has a partial temporal inconsistency 
denoted by $\incp(\Phi)$ iff: 
%
\noindent$\exists \phi, \psi \in \CN(\Phi) \text{ s.t.}$

$~~~\phi$=$P(x_1,$\ldots$,x_k,t_1,t_1') \text{, } \psi$=$\neg P(x_1,$\ldots$,x_k,t_2,t_2') \text{ and } {\TI(t_1,t_1')\cap\TI(t_2,t_2') \neq \emptyset.} $

$~~$Otherwise $\neg \incp(\Phi)$ is true.  

    -- $\Phi$ has a total temporal inconsistency denoted by $\incf(\Phi)$ iff: 
    %
\noindent$\exists \phi, \psi \in \CN(\Phi) \text{ s.t. }$

$~~~ \phi = P(x_1,$\ldots$,x_k,t_1,t_1'), \psi = \neg P(x_1,$\ldots$,x_k,t_2,t_2') \text{ and } (\TI(t_1,t_1') = \TI(t_2,t_2').$


$~~$Otherwise $\neg \incf(\Phi)$ is true. 

\end{definition}



We now examine the interaction properties
between 
$\conp, \conf, \incp$ and $\incf$, 
such as complementarity, subsumption and inclusion \footnote{Proofs are available on \href{https://tinyurl.com/PRICAI22}{https://tinyurl.com/PRICAI22}.}. 



\begin{definition} [Complementarity \& Subsumption]
$\forall \Phi \subseteq \tlogic, \forall \text{ relation } r_1, r_2 \text{ if:} $

-- $r_1(\Phi) \leftrightarrow \neg r_2(\Phi) $ then $r_1$ and $r_2$ are complementary.

-- $r_1(\Phi) \rightarrow r_2(\Phi)$  then $r_1$ subsume $r_2$. 
\end{definition}


The next two propositions show that firstly $\conf$ and $\incp$ 
are complementary; secondly different subsumption relations exist between the temporal consistencies. 

\begin{proposition} \label{prop:dual} 
\textbf{(Complementarity: temporal consistencies)}
For any $\Phi \subseteq \tlogic$: \\ 
$~~~~~~~~~~~~~~~~~~~~~~~~\neg \conf(\Phi)  \leftrightarrow \incp(\Phi) ~\text{and}~\conf(\Phi)  \leftrightarrow \neg \incp(\Phi)$.

\end{proposition}



\begin{proposition} \label{pro:subsum}
\textbf{(Subsumption: temporal consisencies)} 
For any $\Phi \subseteq \tlogic$:\\
$~~~~~~~~~~~~~~~~~~~~~~~~~~~~\conp(\Phi) \rightarrow \neg \incf(\Phi), ~~\incf(\Phi) \rightarrow \neg \conp(\Phi),$\\
$~~~~~~~~~~~~~~~~~~~~~~~~\neg \conp(\Phi) \rightarrow \incp(\Phi), \text{ and}~~\neg \incp(\Phi) \rightarrow \conp(\Phi).$






\end{proposition}

In the following, for temporal consistency and inconsistency relations, we denote by $\{r\} = \{\Phi \subseteq$ $\tlogic \mid r(\Phi) \}$ their set of sets of formulae respecting their condition, where $r \in \{\conp, \conf, \incp, \incf, \neg \conp, \neg \conf, \neg \incp, \neg \incf\}$.

\begin{definition} [Inclusion]\label{prop:implsub}
Let two relations of temporal consistency $r_1, r_2 \in \{\conp, \newline  \conf, \incp, $ $\incf, \neg \conp, \neg \conf, \neg \incp, \neg \incf\}$, $r_1$ is considered included in $r_2$ if:\\
$~~~~~~~~~~~~~~~~~~~~~~~~~~~~~\{r_1\} \subseteq \{r_2\} \text{ iff } \forall \Phi \subseteq \tlogic, r_1(\Phi) \rightarrow r_2(\Phi)$.
\end{definition}

Some inclusions of temporal consistency relations may be defined between the sets of sets of formulae respecting them.
\begin{proposition} \label{prop:inclu}\textbf{(Inclusion: temporal consistencies)}\\
$~~~~~~~~~~~~~~~~~~~~~~~~~~~~~~~~~~~\{\conf\} = \{\neg \incp\} \subseteq \{\conp\} \subseteq \{\neg \incf\}$\\
$~~~~~~~~~~~~~~~~~~~~~~~~~~~~~~~~~~~\{\incf\} \subseteq \{\neg \conp\} \subseteq \{\incp\} = \{\neg \conf\}$
\end{proposition}

\begin{figure}[t]
    \centering
    \includegraphics[width=.9\linewidth]{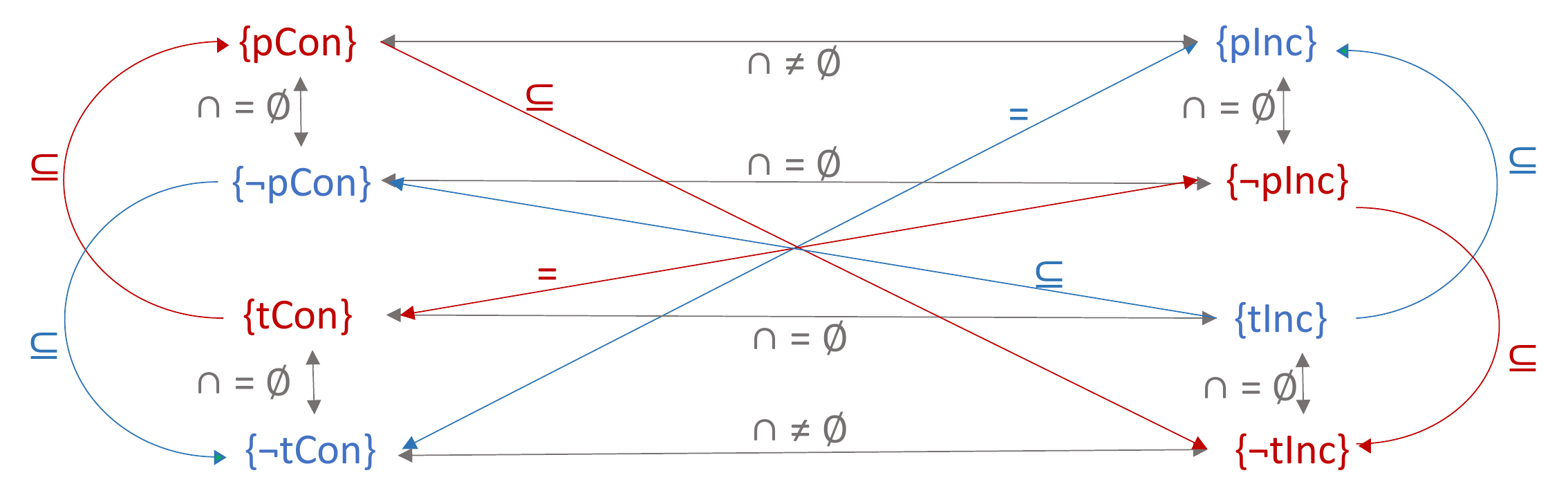}
    \caption{Links between Consistency and Inconsistency}
    \label{fig:Con_Inc}
\end{figure}

Links between different relations of temporal consistency  are illustrated on Figure~\ref{fig:Con_Inc}.

\subsection{Temporal Parametric Semantics}
\label{sec:TPS}

To avoid defining several different semantics, we decompose the construction of semantics and identify three steps. Then, we propose the definition of \textit{Temporal Parametric Semantics}, relying on the combination of three functions: 
i) a \textit{validation} function $\Delta$ of instantiations integrating various consistency relations, 
ii) a \textit{selecting} function $\bigsig$ able to modify the weight of the formulae of an instantiation and
iii) an \textit{aggregate} function $\Theta$ 
returning the final strength.

\begin{definition}[Temporal Parametric Semantics] \label{def:TPS}
A \emph{temporal parametric semantics} is a tuple $\mathtt{TPS} = \tuple{\Delta,\bigsig,\Theta} \in \mathtt{Sem}$, \textit{s.t.}:

        -- $\Delta : \mathtt{TMLN}^* \rightarrow \{0,1\}$,
        
        -- $\bigsig : \mathtt{TMLN}^* \rightarrow \bigcup^{+ \infty}_{k=0} [0,1]^k$,
        
        -- $\Theta  : \bigcup^{+ \infty}_{k=0} [0,1]^k  \rightarrow [0, + \infty[$, 

\noindent For any $\bM \in \mathtt{TMLN}$, $I \subseteq \MI(\bM)$, the strength of a temporal parametric semantics $\mathtt{TPS}  = \tuple{\Delta,\bigsig,\Theta}$ is computed by:
$\mathtt{TPS}(I) = \Delta(I) \cdot \Theta\Big(\bigsig(I)\Big). $

	
	

\end{definition}

We propose below key properties that must be satisfied by each of the three functions $\Delta, \bigsig, \Theta$ of a \texttt{TPS}.
Those properties constrain the range of functions to be considered, and discard those exhibiting undesired behaviours. 


\begin{definition}
\label{def:wellTPS}
A temporal parametric semantics $\mathtt{TPS} = \tuple{\Delta,\bigsig,\Theta}$ is \emph{well-behaved} according to a temporal consistency relation $\mathtt{Con}$ iff the following conditions hold:
\begin{enumerate}
\item[$\Delta$-] 
		\begin{description}
			
 			\item [(a)] $\Delta(I) = 1$ if $\mathtt{Con}(\TF(I))$ is true (\ie $I$ is temporally consistent w.r.t. $\mathtt{Con}$).
 			
		\end{description}
		
\item[$\Theta$-]
		\begin{description} 
			
			\item [(a)] $\Theta() = 0$.
			\item [(b)] $\Theta(w) = w$. 
			\item [(c)] $\Theta$ is symmetric.
			\item [(d)] $\Theta(w_1, $\ldots$, w_k) = \Theta(w_1, $\ldots$, w_k, 0)$.  
			\item [(e)] $\Theta(w_1, $\ldots$, w_k, y) \leq \Theta(w_1, $\ldots$, w_k,z)$ 
			            if $y \leq z$.

		\end{description}
		
\item[$\bigsig$-]
		\begin{description} 
			\item [(a)] $\bigsig() = ()$. 
 			\item [(b)] $\bigsig(\{(\phi_1,w_1),$\ldots$,(\phi_k,w_k)\})= (w'_1, $\ldots$,w'_n)$ such that if $k \geq 1$ then $n \geq 1$. 
 			\item [(c)] $\bigsig(\{(\phi_1,w_1),$\ldots$,(\phi_k,w_k),(\phi_{k+1},0)\})= (\bigsig(\{(\phi_1,w_1),$\ldots$,(\phi_k,w_k)\}),0)$,\\
 			 if $\bigtau(\TF(\{\phi_1,$\ldots$,\phi_k\})) \not\vdash \bigtau(\{\phi\})$.
 			 \item [(d)] $\bigsig(\{(\phi_1,w_1),$\ldots$,(\phi_k,w_k)\})\subset \bigsig(\{(\phi_1,w_1),$\ldots$,(\phi_k,w_k),(\phi_{k+1},w_{k+1})\})$\\ if
 			$\phi_{k+1}$ is a ground temporal formula, 
 			$\bigtau(\TF(\{\phi_1,$\ldots$,\phi_k\})) \not\vdash \bigtau(\{\phi_{k+1}\})$ and
 			$\mathtt{Con}(\TF(\{(\phi_1,w_1),$\ldots$,(\phi_k,w_k),(\phi_{k+1},w_{k+1})\}))$.
 			\item [(e)] $\Theta\Big(\bigsig(\{(\phi_1,w_1),$\ldots$,(\phi_k,w_k)\})\Big)\leq \Theta\Big(\bigsig(\{(\phi_1,w_1),$\ldots$,(\phi_k,w_k),$\\$(\phi_{k+1},w_{k+1})\})\Big)$ 
 			if 
 			$\bigtau(\TF(\{\phi_1,$\ldots$,\phi_k\})) \not\vdash \bigtau(\{\phi_{k+1}\})$ 
 			and\\ $\mathtt{Con}(\TF(\{(\phi_1,w_1),$\ldots$,(\phi_k,w_k),(\phi_{k+1},w_{k+1})\}))$.
		\end{description}
		
\end{enumerate}
We also say that $\Delta$ is \textit{well-behaved} according to $\mathtt{Con}$, and $\Theta, \bigsig$ are \textit{well-behaved}. 
\end{definition}

\begin{theorem}\label{theo:satprin}
Any temporal parametric semantics well-behaved w.r.t a temporal consistency relation $\mathtt{Con}$ satisfies the principles from Section~\ref{sec:principles}: Temporal Neutrality, Consistency Monotony and Invariant Consistent Facts (the last two according to $\mathtt{Con}$).
\end{theorem}



Once temporal consistency relations are defined, we may enhance semantics for MAP inference with temporal validation functions.
One TMLN instantiation can be valid or not according to different criteria (\textit{i.e.,} accept an instantiation).



\begin{definition}[Temporal Consistency Validation Function] \label{def:tcvf}
Let $\bM \in \mathtt{TMLN}$, an instantiation $I \subseteq \MI(\bM)$ and $x \in \{\conp, \conf, \incp, \incf\}$.
We define $\Delta_x : \mathtt{TMLN^*} \rightarrow \{0,1\}$, a temporal consistency validation function according to $x$, such that: 

\small{
\begin{tabular}{l}
\setlength\tabcolsep{0pt}
$\begin{array}{l} 
\Delta_{\conp}(I) =  
\left\{\begin{array}{ll}
1 & \text{ if $\conp(\TF(I))$ }  \\
0 & \text{ if $\neg \conp(\TF(I))$}
\end{array}\right.\end{array}$
$\begin{array}{l} 
\Delta_{\conf}(I) =  
\left\{\begin{array}{ll}
1 & \text{ if $\conf(\TF(I))$ }  \\
0 & \text{ if $\neg \conf(\TF(I))$}
\end{array}\right.\end{array}$
\\

\setlength\tabcolsep{0pt}
    $\begin{array}{l} 
\Delta_{\incp}(I) =  
\left\{\begin{array}{ll}
0 &
 \text{ if $\incp(\TF(I))$ }  \\
1 &
 \text{ if $\neg \incp(\TF(I))$}
\end{array}\right.\end{array}$

    $\begin{array}{l} 
\Delta_{\incf}(I) =  
\left\{\begin{array}{ll}
0 & \text{ if $\incf(\TF(I))$ }  \\
1 & \text{ if $\neg \incf(\TF(I))$}
\end{array}\right.\end{array}$\\
\end{tabular}
}

\end{definition}

\begin{corollary}
For any $I \subseteq \mathtt{TMLN}^*$, $\Delta_{\conf}(I) = \Delta_{\incp}(I)$.
\end{corollary}

\begin{corollary}
Let $x \in \{\conp, \conf, \incp, \incf\}$, each $\Delta_x$ is well-behaved.
\end{corollary}

We show next that we can order the value of the $\Delta_x$ for any instantiation.

\begin{proposition} \label{prop:ordre}
Let $\bM \in \mathtt{TMLN}$ and 
$\Delta_x$ a temporal consistency validation function such that
 $x \in \{\conp, \conf, \incp, \incf\}$.
For any instantiation $I \subseteq \MI(\bM)$:\\
$~~~~~~~~~~~~~~~~~~~~~~~~~~~~~~~\Delta_{\conf}(I) = \Delta_{\incp}(I) \leq \Delta_{\conp}(I) \leq \Delta_{\incf}(I)$
\end{proposition}

Theorem~\ref{theo:ordrew} shows that the strength of the temporal MAP inferences with $\bigsig$ and $\Theta$ on any TMLN is ranked according to the temporal consistency validation functions $\Delta_x$.
\begin{theorem} \label{theo:ordrew}
Let $\bM \in \mathtt{TMLN}$, for any $\bigsig$ and $\Theta$, we denote by: 

    -- $\mathtt{TPS}_{\conf} = \tuple{\Delta_{\conf},\bigsig,\Theta}$, 
    $\mathtt{TPS}_{\incp} = \tuple{\Delta_{\incp},\bigsig,\Theta}$, 
    
    -- $\mathtt{TPS}_{\conp} = \tuple{\Delta_{\conp},\bigsig,\Theta}$, 
    $\mathtt{TPS}_{\incf} = \tuple{\Delta_{\incf},\bigsig,\Theta}$.\\
Hence: 
$\forall I_{\conf} \in \mathtt{map}(\bM,\mathtt{TPS}_{\conf})$, $\forall I_{\incp} \in \mathtt{map}(\bM,\mathtt{TPS}_{\incp})$,
$\forall I_{\conp} \in \linebreak \mathtt{map}(\bM,\mathtt{TPS}_{\conp})$,
$\forall I_{\incf} \in \mathtt{map}(\bM,\mathtt{TPS}_{\incf})$, 

\vspace{0.1cm}
$~~~~~~~~~\mathtt{TPS}_{\conf}(I_{\conf}) = 
\mathtt{TPS}_{\incp}(I_{\incp}) \leq\mathtt{TPS}_{\conp}(I_{\conp}) \leq
\mathtt{TPS}_{\incf}(I_{\incf}).$
\end{theorem}




We study different instances of the aggregate function, using different sums.
This type of parameters will determine the strength of an instantiation, in various ways.

\begin{definition}[Aggregate Functions] \label{def:agg}
Let $\{w_1, $\ldots$, w_n\}$ such that $n \in [0, +\infty[$ 
and $\forall i \in [0,n]$, $w_i \in [0,1]$.

    -- $\Theta_{sum}(w_1, $\ldots$, w_n) = \sum^{n}\limits_{i = 1} w_i$, 
    if $n=0$ then  ${\Theta_{sum}() = 0.}$
    
    -- $\Theta_{sum,\alpha}(w_1, $\ldots$, w_n) = \left(\sum^{n}\limits_{i = 1} (w_i)^{\alpha}\right)^{\frac{1}{\alpha}}$ s.t. $\alpha \geq 1$, 
    if $n=0$ then $\Theta_{sum,\alpha}() = 0$.
    
    -- $\Theta_{psum}(w_1, $\ldots$, w_n) = w_1 \ominus $\ldots$ \ominus w_n$, where $w_1 \ominus w_2 = w_1 + w_2 - w_1 \cdot w_2$, 
    
    $~~$ if $n=0$ then $\Theta_{psum}() = 0$ and if $n=1$ then $\Theta_{psum}(w) = w$.
\end{definition}

Those aggregate functions target different kinds of semantics. For instance, $\Theta_{sum,\alpha}$ emphasises strong weights for inference, while $\Theta_{psum}$ targets more controversial instantiations (mixed weights).

\begin{proposition} \label{prop:ThetaWB}
The three functions $\Theta_{sum}$, $\Theta_{sum, \alpha}$ and $\Theta_{psum}$ are well-behaved.
\end{proposition}

We propose below a selective function $\bigsig_{id}$
which returns all the weights, more complex functions selecting weights with a threshold ($\bigsig_{thresh,\alpha}$) and another one, $\bigsig_{rule}$, setting the weight of ground rules to 0 when one of their premises is not deductible (\textit{i.e.}, the rule is unusable).
The latter is interesting to say if unground rules (no deductible premises -- \texttt{prem \& imp}) must be kept or not in any instantiation.

\begin{definition}[Selective Functions] \label{def:sel}
Let $\bM \in \mathtt{TMLN}$, $\{(\phi_1,w_1),$\ldots$,(\phi_n,w_n)\} \subseteq \MI(\bM)$:
\begin{itemize}
    \item $\bigsig_{id}(\{(\phi_1,w_1),$\ldots$,(\phi_n,w_n)\}) = (w_1, $\ldots$, w_n)$
    
    \item $\bigsig_{thresh,\alpha}(\{(\phi_1,w_1),$\ldots$,(\phi_n,w_n)\}) = (\mathtt{max}(w_1 - \alpha, 0), $\ldots$, \mathtt{max}(w_n - \alpha, 0))$ s.t. $\alpha \in [0,1[$
    
    \item \begin{itemize}
        \item let $\phi = (\psi_1 \wedge $\ldots$ \wedge \psi_k) \rightarrow \psi_{k+1}$ a rule, \\ $\mathtt{prem}(\phi) = \{\psi_1, $\ldots$, \psi_k\}$.
        \vspace{-.2cm}
        \item $\mathtt{imp}((\phi,w),\{(\phi_1,w_1), $\ldots$, (\phi_n,w_n)\}) = $ 
        $\begin{array}{l} 
                  \left\{\begin{array}{ll}
                 0 & \text{if $\phi$ is a ground rule}\\
                 &\text{s.t. $\exists \psi_i \in \mathtt{prem}(\phi)$ }\\
                  & \text{s.t. $\psi_i \notin \CN(\{\phi_1, $\ldots$,\phi_n\})$ }  \\
                 w & \text{ otherwise}
                 \end{array}\right.\end{array}$
                 
        \item $\bigsig_{rule}(\{(\phi_1,w_1),$\ldots$,(\phi_n,w_n)\}) = \Big(\mathtt{imp}((\phi_1,w_1),\{(\phi_2,w_2), $\ldots$,$ \\ $(\phi_n,w_n)\}), $\ldots$, \mathtt{imp}((\phi_n,w_n),\{(\phi_1,w_1), $\ldots$, (\phi_{n-1},w_{n-1})\}) \Big) $ 
    \end{itemize}
    

        
        

    
\end{itemize}

\end{definition}

\begin{proposition} \label{prop:sigmaWB}
The three functions $\bigsig_{id}$, $\bigsig_{thresh,\alpha}$ and $\bigsig_{rule}$ are well-behaved.
\end{proposition}

In~\cite{chekol2017marrying}, the MAP inference uses a semantics working on \texttt{Herbrand} models (containing temporal formulae, included in $\tlogic$) and built from uncertain ($w < 1$) and certain ($w = 1$) temporal facts and with a set of TMLN certain rules. 
This semantics also determines the temporal inconsistency by a classical consistency (if there is no formula $\varphi$ such that $\Phi \vdash \varphi$ and $\Phi \vdash \lnot \varphi$) and by summing all the weights of facts in the instantiation. 
Therefore, for TMLNs without any uncertain rule, the MAP inference will return the same instantiations as ours, using the temporal parametric semantics $ \tuple{\Delta_{\incf},\bigsig_{id},\Theta_{sum}}$.
%
Our Temporal MAP inference generalises their work.

\subsection{Example of Reasoning on TMLN}


\begin{exs}{ex:running}
Table~\ref{tab:exTPS} 
illustrates various $\mathtt{TPS}$ leading to different instantiations and possible inferences from Example~\ref{ex:running}.
The last column focuses on the \textit{PeasantFamily} information which has not been proven yet by Historians.
Another interesting analyse is about "\textit{did Nicole Oresme studied at the College of Navarre and when?}".  

\begin{table*}[t]
    \centering
    \scriptsize
    \begin{tabular}{|c|l|r|}
    \hline
    \textbf{Temporal Parametric Semantics} & \multicolumn{1}{c|}{\textbf{Temporal MAP Inferences}} & \multicolumn{1}{c|}{\textbf{Example of Conclusion}} \\
    \hline
    $\tuple{\Delta_{\conf},\bigsig_{id},\Theta_{sum}}$ & $\{\{ F_6, GR_{11},GR_{12}, GR_2\}\}$ &  $(\neg PeasantFamily(NO,t_{min},t_{max}),0.8)$ \\
    $\tuple{\Delta_{\conp},\bigsig_{id},\Theta_{sum}}$ & $\{\{ F_4, F_6, GR_{12}, GR_2\}\}$  & $(\neg PeasantFamily(NO,t_{min},t_{max}),0.8)$ \\
    $\tuple{\Delta_{\incf},\bigsig_{id},\Theta_{sum}}$ & $\{\{F_4, F_5, F_6, GR_{11}, GR_{12} \}\}$ & $(PeasantFamily(NO,t_{min},t_{max}),0.5)$ \\
    \hline
    \hline
    $\tuple{\Delta_{\conf},\bigsig_{id},\Theta_{sum,2}}$ &$\{\{ F_6, GR_{11},GR_{12}, GR_2\}\}$ &  $(\neg PeasantFamily(NO,t_{min},t_{max}),0.8)$ \\
    $\tuple{\Delta_{\conp},\bigsig_{id},\Theta_{sum,2}}$ & $\{\{ F_4, F_6, GR_{12}, GR_2\}\}$  & $(\neg PeasantFamily(NO,t_{min},t_{max}),0.8)$ \\
    $\tuple{\Delta_{\incf},\bigsig_{id},\Theta_{sum,2}}$ & $\{\{F_4, F_5, F_6, GR_2\},$  & $(\neg PeasantFamily(NO,t_{min},t_{max}),0.8)$ \\
     & \hspace{.5cm}$ \{F_5, F_6, GR_{11}, GR_2\}\}$ & \\
    \hline
    \hline
    $\tuple{\Delta_{\conf},\bigsig_{rule},\Theta_{sum}}$ & $\{\{ F_4, F_5, GR_{11}, GR_{12} \}\}$  &  $(PeasantFamily(NO,t_{min},t_{max}),0.5)$ \\
    $\tuple{\Delta_{\conp},\bigsig_{rule},\Theta_{sum}}$ & $\{\{ F_4, F_6, GR_2\}\}$ & $(\neg PeasantFamily(NO,t_{min},t_{max}), 0.8)$ \\
    $\tuple{\Delta_{\incf},\bigsig_{rule},\Theta_{sum}}$ & $\{\{F_4, F_5,F_6, GR_{11}, GR_{12}\}\}$ & $(PeasantFamily(NO,t_{min},t_{max}),0.5)$ \\
    \hline
    \hline
    $\tuple{\Delta_{\conf},\bigsig_{rule},\Theta_{sum,2}}$ & $\{\{F_6, GR_2\}\}$  &  $(\neg PeasantFamily(NO,t_{min},t_{max}), 0.8)$ \\
    $\tuple{\Delta_{\conp},\bigsig_{rule},\Theta_{sum,2}}$ & $\{\{ F_4, F_6, GR_2\}\}$  & $(\neg PeasantFamily(NO,t_{min},t_{max}), 0.8)$ \\
    $\tuple{\Delta_{\incf},\bigsig_{rule},\Theta_{sum,2}}$ & $\{\{ F_4, F_5, F_6, GR_2\}\}$  & $(\neg PeasantFamily(NO,t_{min},t_{max}), 0.8)$ \\
    \hline
    \end{tabular}
    \caption{TPS example: in each instantiation we omit $F_1,F_2,F_3$ and formulae with a 0 weight.}
    \label{tab:exTPS}
\end{table*}


$\Theta_{sum,2}$ keeps highly weighted facts and rules. Instantiations insist on the conflict between $GR_{11} / GR_{12}$ and $GR_2$ (\textit{PeasantFamily} or not), keeping those which maximise the strength.
More precisely, these ground rules are temporally inconsistent depending on the set of facts in the instantiation. Indeed, if ground rules' premises are not present in the instantiation, then they have no conflict, as it appears in the $\mathtt{TPS}$ using $\bigsig_{id}$.


$\bigsig_{rule}$ gives an interesting vision on the MAP inference result, since it keeps (\ie \linebreak $w > 0$) only useful ground rules, \ie\,rules  having deductible premises in the instantiation. Hence, 
$\bigsig_{rule}$ never give importance simultaneously to $GR_{11} / GR_{12}$ with $GR_2$.


Eventually, notice that even if $PeasantFamily(NO,t_{min},t_{max})$ has a weight of $0.5$ and its negation $0.8$, or even if in this example we have more inference on the negation, we cannot conclude that $\neg PeasantFamily(NO,t_{min},t_{max})$ is more likely.
For instance, $\tuple{\Delta_{\incf},\bigsig_{id},\Theta_{sum}}$ corresponding to Chekol's semantics, advocates that Nicole Oresme was a Peasant while the one with $\Delta_{\conp}$ (the partial temporal consistency) infers the opposite.
The parametric choice of our approach would allow the historian deciding which inference corresponds to their own reasoning.


\end{exs}

\section{Conclusion}

Reasoning on Knowledge Graphs have recently been approached with MLNs, to find the most probable state of the world. However, they were limited to a strict temporal inconsistency.
We extend UTKGs with a fully-formalized model of TMLN, capable of combining consistency and inconsistency for facts and rules, allowing to reason with the MAP inference using new principles and relations between partial and total aspects of temporal (in-)consistency. 
Moreover, we introduce temporal parametric semantics which offer flexibility to tailor semantics reasoning to one's needs.

As a perspective, we first wish to extend this work to existential rules in order to capture more real data (see Section~\ref{sec:instantiation}). 
Secondly, some relationships between formulae in a TMLN are not yet exploited and could be represented with argumentation graphs (\textit{e.g.,} the notion of support \cite{cohen2014survey} or similarity between pieces of information \cite{AmgoudD18,AmgoudD21}), to enhance the weight of inferred knowledge. 
Thirdly, as we can see in Table \ref{tab:exTPS}, each parameter of the $\mathtt{TPS}$ have an impact on the temporal MAP inferences and its possible conclusions. It will be useful to investigate some properties to describe the strategies and specific behaviours of the parametrisation functions.
Finally, implementing our results and confronting them with real data will likely prove very useful, at least to generate new research hypotheses in historical science.

\clearpage

\bibliographystyle{splncs04}
\bibliography{biblio}

\clearpage

\begin{appendix}

\textbf{Supplementary material for\\  Parameterisation of Reasoning on \\ Temporal Markov Logic Networks}

\section{Proofs}\label{sec:proofs}

\begin{proof} [Proposition \ref{prop:dual}] 
Let $\Phi \subseteq \tlogic$ such that $\conf(\Phi)$, i.e. from Definition \ref{def:ticon}:
$$\forall \phi, \psi \in \CN(\Phi) \text{ s.t. } \phi = P(x_1,\cdots,x_k,t_1,t_1') \text{ and }$$
$$ \psi = \neg P(x_1,\cdots,x_k,t_2,t_2') \text{, } (\TI(t_1,t_1') \cap \TI(t_2,t_2') = \emptyset) $$

then its negation $\neg \conf(\Phi)$ is equivalent to:
$$\neg (\neg \exists \phi, \psi \in \CN(\Phi) \text{ s.t. } \phi = P(x_1,\cdots,x_k,t_1,t_1') \text{ and }$$
$$ \psi = \neg P(x_1,\cdots,x_k,t_2,t_2') \text{ and } (\TI(t_1,t_1') \cap \TI(t_2,t_2') \neq \emptyset)) $$


Hence $\neg \conf(\Phi)$ is equivalent to $\incp(\Phi)$:
$$\exists \phi, \psi \in \CN(\Phi) \text{ s.t. } \phi = P(x_1,\cdots,x_k,t_1,t_1') \text{ and }$$
$$\psi = \neg P(x_1,\cdots,x_k,t_2,t_2') \text{ and } (\TI(t_1,t_1') \cap \TI(t_2,t_2') \neq \emptyset) $$

Moreover its negation, $\neg \incp(\Phi)$ is equivalent to:
$$\neg (\neg \forall \phi, \psi \in \CN(\Phi) \text{ s.t. } \phi = P(x_1,\cdots,x_k,t_1,t_1') \text{ and }$$
$$ \psi = \neg P(x_1,\cdots,x_k,t_2,t_2') \text{, } (\TI(t_1,t_1') \cap \TI(t_2,t_2') = \emptyset))$$
Therefore $\neg \incp(\Phi)$ is equivalent to  $\conf(\Phi)$.
\end{proof}

\vspace{0.5cm}


\begin{proof} [Proposition \ref{pro:subsum}]
Let $\Phi \subseteq \tlogic$, from Definition \ref{def:ticon}:
\begin{enumerate}
    \item $\conp(\Phi) \rightarrow \neg \incf(\Phi)$: 

\begin{itemize}
    \item $\neg \incf(\Phi)$ iff,
$$\neg \exists \phi, \psi \in \CN(\Phi) \text{ s.t. } \phi = P(x_1,\cdots,x_k,t_1,t_1'),$$
$$\psi = \neg P(x_1,\cdots,x_k,t_2,t_2') \text{ and } (\TI(t_1,t_1') = \TI(t_2,t_2')) $$
    
    \item $\conp(\Phi)$ iff 
$$\forall \phi, \psi \in \CN(\Phi) \text{ s.t. } \phi = P(x_1,\cdots,x_k,t_1,t_1') \text{ and }$$
$$ \psi = \neg P(x_1,\cdots,x_k,t_2,t_2') \text{, } (\TI(t_1,t_1') \setminus \TI(t_2,t_2') \neq \emptyset) \wedge (\TI(t_2,t_2') \setminus \TI(t_1,t_1') \neq \emptyset) $$

     which is equivalent to:
$$\neg \exists \phi, \psi \in \CN(\Phi) \text{ s.t. } \phi = P(x_1,\cdots,x_k,t_1,t_1') \text{ and }$$
$$ \psi = \neg P(x_1,\cdots,x_k,t_2,t_2') \text{ and } $$
$$(\TI(t_1,t_1') \setminus \TI(t_2,t_2') = \emptyset) \vee (\TI(t_2,t_2') \setminus \TI(t_1,t_1') = \emptyset)$$

    \item If there not exists $\TI(t_1,t_1') \setminus \TI(t_2,t_2') = \emptyset \text{ or } \TI(t_2,t_2') \setminus \TI(t_1,t_1') = \emptyset$ then there not exists $\TI(t_1,t_1') \setminus \TI(t_2,t_2') = \emptyset \text{ and } \TI(t_2,t_2') \setminus \TI(t_1,t_1') = \emptyset$ (i.e. $\TI(t_1,t_1') = \TI(t_2,t_2')$), therefore  $\conp(\Phi) \rightarrow \neg \incf(\Phi) $.\\
\end{itemize}


\item $\incf(\Phi) \rightarrow \neg \conp(\Phi)$: 

\begin{itemize}
    \item $\neg \conp(\Phi)$ iff 
$$\neg(\neg \exists \phi, \psi \in \CN(\Phi) \text{ s.t. } \phi = P(x_1,\cdots,x_k,t_1,t_1') \text{ and }$$
$$ \psi = \neg P(x_1,\cdots,x_k,t_2,t_2') \text{ and }$$
$$(\TI(t_1,t_1') \setminus \TI(t_2,t_2') = \emptyset) \vee (\TI(t_2,t_2') \setminus \TI(t_1,t_1') = \emptyset))$$
i.e.,
$$\exists \phi, \psi \in \CN(\Phi) \text{ s.t. } \phi = P(x_1,\cdots,x_k,t_1,t_1') \text{ and }$$
$$ \psi = \neg P(x_1,\cdots,x_k,t_2,t_2') \text{ and }$$
$$(\TI(t_1,t_1') \setminus \TI(t_2,t_2') = \emptyset) \vee (\TI(t_2,t_2') \setminus \TI(t_1,t_1') = \emptyset)$$

\item 
 $\incf(\Phi)$ iff
     $$\exists \phi, \psi \in \CN(\Phi) \text{ s.t. } \phi = P(x_1,\cdots,x_k,t_1,t_1'),$$
$$\psi = \neg P(x_1,\cdots,x_k,t_2,t_2') \text{ and } (\TI(t_1,t_1') = \TI(t_2,t_2')) $$

\item 
If there exists $\TI(t_1,t_1') \setminus \TI(t_2,t_2') = \emptyset \text{ and } \TI(t_2,t_2') \setminus \TI(t_1,t_1') = \emptyset$ ($\TI(t_1,t_1') = \TI(t_2,t_2')$), then there exists $\TI(t_1,t_1') \setminus \TI(t_2,t_2') = \emptyset \text{ or } \TI(t_2,t_2') \setminus \TI(t_1,t_1') = \emptyset$, therefore  $\incf(\Phi) \rightarrow \neg \conp(\Phi)$.\\
\end{itemize}

\item $\neg \conp(\Phi) \rightarrow \incp(\Phi)  $: 
\begin{itemize}
    \item $\incp(\Phi)$ iff
$$\exists \phi, \psi \in \CN(\Phi) \text{ s.t. } \phi = P(x_1,\cdots,x_k,t_1,t_1') \text{, }$$
$$\psi = \neg P(x_1,\cdots,x_k,t_2,t_2') \text{ and } (\TI(t_1,t_1') \cap \TI(t_2,t_2') \neq \emptyset) $$

    \item If there exists $\TI(t_1,t_1') \setminus \TI(t_2,t_2') = \emptyset \text{ or } \TI(t_2,t_2') \setminus \TI(t_1,t_1') = \emptyset$ then there exists $\TI(t_1,t_1') \cap \TI(t_2,t_2') \neq \emptyset$, therefore  $\neg \conp(\Phi) \rightarrow \incp(\Phi)  $.\\
\end{itemize}

\item $\neg \incp(\Phi) \rightarrow \conp(\Phi)  $: 
\begin{itemize}
\item $\neg \incp(\Phi)$ is equivalent to:
$$\neg( \neg \forall \phi, \psi \in \CN(\Phi) \text{ s.t. } \phi = P(x_1,\cdots,x_k,t_1,t_1') \text{ and }$$
$$ \psi = \neg P(x_1,\cdots,x_k,t_2,t_2') \text{, } (\TI(t_1,t_1') \cap \TI(t_2,t_2') = \emptyset) $$

\item Therefore, for any $\TI(t_1,t_1'), \TI(t_2,t_2')$, if $\TI(t_1,t_1') \cap \TI(t_2,t_2') = \emptyset $ then $\TI(t_1,t_1') \setminus \TI(t_2,t_2') \neq \emptyset \text{ and } \TI(t_2,t_2') \setminus \TI(t_1,t_1') \neq \emptyset$, therefore  $\neg \incp(\Phi) \rightarrow \conp(\Phi)  $.\\
\end{itemize}

\end{enumerate}
\end{proof}


\vspace{0.1cm}

\begin{proof} [Proposition \ref{prop:inclu}] 
From Proposition \ref{prop:dual} and \ref{pro:subsum}, we know that for any $\Phi \subseteq \tlogic$:
$$\conf(\Phi) \leftrightarrow \neg \incp(\Phi) \rightarrow \conp(\Phi) \rightarrow \neg \incf(\Phi)$$
$$\incf \rightarrow \neg \conp(\Phi) \rightarrow \incp(\Phi) \leftrightarrow \neg \conf(\Phi)$$

Therefore from Proposition \ref{prop:implsub}: 
$$\{\conf\} = \{\neg \incp\} \subseteq \{\conp\} \subseteq \{\neg \incf\}$$
$$\{\incf\} \subseteq \{\neg \conp\} \subseteq \{\incp\} = \{\neg \conf\}$$


\end{proof}

\vspace{0.5cm}

\begin{proof} [Theorem \ref{theo:satprin}] {\color{white}.}
\begin{itemize}
    \item Temporal Neutrality.\\
    Let $\bM \in \mathtt{TMLN}$, $\bM' = \bM \cup \{(\phi,w)\}$ where $(\phi,w)$ is a weighted temporal formula such that:
    \begin{itemize}
        \item $\bigtau(\TF(\bM)) \not\vdash \bigtau(\{\phi\})$ 
        \item $w = 0$.
    \end{itemize}
    
    Given that $(\phi,w = 0)$, any new ground formula in the instantiation of $\bM'$ has also a zero weight: $\forall (\psi,w) \in \MI(\bM')\setminus \MI(\bM)$, $w = 0$ (because the instantiation take the minimum weight of the formulae building the ground formula).
    
    If a temporal parametric semantics $\mathtt{TPS} = \tuple{\Delta,\bigsig,\Theta}$ is \emph{well-behaved} according to a temporal consistency relation $\mathtt{Con}$, then 
    from Definition \ref{def:wellTPS}: 
    \begin{itemize}
        \item condition 2.(c): $\Theta(w_1, \cdots, w_k) = \Theta(w_1, \cdots, w_k, 0)$. 
        \item condition 3.(c): $\bigsig(\{(\phi_1,w_1),\cdots,(\phi_k,w_k),$ \\
        $(\phi_{k+1},0)\})
 			= (\bigsig(\{(\phi_1,w_1),\cdots,(\phi_k,w_k)\}),0)$,\\
 			 if $\bigtau(\TF(\{\phi_1,\cdots,\phi_k\})) \not\vdash \bigtau(\{\phi\})$.
    \end{itemize}
    
    Therefore, any new ground formula will have no impact in any instantiation, and so in any MAP inference:
    $\forall I \in \mathtt{map}(\bM,\mathtt{TPS})$ and $\forall I' \in \mathtt{map}(\bM',\mathtt{TPS})$, $\mathtt{TPS}(I) = \mathtt{TPS}(I')$.

    \item Consistency Monotony.\\ 
    Let a well-behaved temporal semantics $\mathtt{TPS} = \tuple{\Delta,\bigsig,\Theta}$ according to a temporal consistency relation $\mathtt{Con}$, $\bM \in \mathtt{TMLN}$, $\bM' = \bM \cup \{(\phi,w)\}$ where $(\phi,w)$ is a weighted temporal formula such that:
    \begin{enumerate}
    \item $\bigtau(\TF(\bM)) \not\vdash \bigtau(\{\phi\})$, and
    \item $\forall I \in \mathtt{map}(\bM,\mathtt{TPS})$, $\mathtt{Con}(\{\phi\} \cup \TF(I))$ is true and $I \subset \MI(\{(\phi,w)\} \cup I)$.
    \end{enumerate}  
    
    From Definition \ref{def:wellTPS}: 
    \begin{itemize}
		\item condition 1.(a): $\Delta(I) = 1$ if $\mathtt{Con}(\TF(I))$ is true.	
        \item condition 3.(e):  $\Theta\Big(\bigsig(\{(\phi_1,w_1),\cdots,(\phi_k,w_k)\})\Big)$\\
 			$\leq \Theta\Big(\bigsig(\{(\phi_1,w_1),\cdots,(\phi_k,w_k),(\phi_{k+1},w_{k+1})\})\Big)$ if 
 			$\bigtau(\TF(\{\phi_1,\cdots,\phi_k\})) \not\vdash \bigtau(\{\phi_{k+1}\})$ 
 			and $\mathtt{Con}(\TF(\{(\phi_1,w_1),\cdots,(\phi_k,w_k),(\phi_{k+1},w_{k+1})\}))$.
    \end{itemize}
    
    Therefore, from item 2 of $(\phi,w)$ we know that any new ground formulae are consistent with each MAP, and with condition 1.(a), we know that the strength will not down to 0 from $\Delta$.
    Given that we have at least one new ground formula in each MAP 
    (condition 2 of Consistency Monotony $I \subset \MI(\{(\phi,w)\} \cup I)$),
    with condition 3.(e) we know that:\\
    $\forall I \in \mathtt{map}(\bM,\mathtt{TPS})$ and $\forall I' \in \mathtt{map}(\bM',\mathtt{TPS})$, 
    $\mathtt{TPS}(I) \leq \mathtt{TPS}(I')$.

    \item Invariant Consistent Facts.\\
    Let  a well-behaved temporal semantics $\mathtt{TPS} = \tuple{\Delta,\bigsig,\Theta}$ according to a temporal consistency relation $\mathtt{Con}$, $\bM \in \mathtt{TMLN}$, $\bM' = \bM \cup \{(\phi,w)\}$ where $(\phi,w)$ is a TMLN fact such that:
    \begin{itemize}
        \item $\bigtau(\TF(\bM)) \not\vdash \bigtau(\{\phi\})$, and
        \item $\forall I \in \mathtt{map}(\bM,\mathtt{TPS})$, $\mathtt{Con}(\{\phi\} \cup \TF(I))$ is true.
    \end{itemize}   
    
     From Definition \ref{def:wellTPS}: 
    \begin{itemize}
		\item condition 1.(a): $\Delta(I) = 1$ if $\mathtt{Con}(\TF(I))$ is true.	
        \item condition 3.(d): $\bigsig(\{(\phi_1,w_1),\cdots,(\phi_k,w_k)\})$\\
 			$\subset \bigsig(\{(\phi_1,w_1),\cdots,(\phi_k,w_k),(\phi_{k+1},w_{k+1})\})$ if
 			$\phi_{k+1}$ is a ground temporal formula, 
 			$\bigtau(\TF(\{\phi_1,\cdots,\phi_k\})) \not\vdash \bigtau(\{\phi_{k+1}\})$ and \linebreak
 			$\mathtt{Con}(\TF(\{(\phi_1,w_1),\cdots,(\phi_k,w_k),(\phi_{k+1},w_{k+1})\}))$.
    \end{itemize}
    
    Given that $(\phi,w)$ is a TMLN fact, then $\phi$ is a ground formula.
    Moreover, from the definition of Temporal MAP Inference, we know that each MAP is maximal for set inclusion ($\mathtt{map}(\bM,S) = \{ I \mid I \in \underset{I ~\subseteq ~\MI(\bM) }{\text{argmax}}~S(I) \text{ and }$
 $\nexists I' \in \underset{I' ~\subseteq ~\MI(\bM) }{\text{argmax}}~S(I') \text{ s.t. } \\ I \subset I'\}$).
    Therefore the new consistent temporal fact $(\phi,w)$ will be add to any previous MAP: 
    $\forall I \in \mathtt{map}(\bM,\mathtt{TPS})$,  $I \cup \{(\phi,w)\} \in \mathtt{map}(\bM',\mathtt{TPS})$. 

\end{itemize}

\end{proof}

\vspace{0.5cm}

\begin{proof} [Proposition \ref{prop:ordre}]. 
From Definition \ref{prop:implsub}: Let two relations of temporal consistency $r_1, r_2 \in \{\conp, \conf, \incp, \incf, \neg \conp, \neg \conf,$ $\neg \incp, \neg \incf\}$, $r_1$ is considered included in $r_2$ if:
$$ \{r_1\} \subseteq \{r_2\} \text{ iff } \forall \Phi \subseteq \mathtt{TS}-\mathtt{FOL}, r_1(\Phi) \rightarrow r_2(\Phi)$$

From Proposition \ref{prop:inclu}:
$$\{\conf\} = \{\neg \incp\} \subseteq \{\conp\} \subseteq \{\neg \incf\}$$
$$\{\incf\} \subseteq \{\neg \conp\} \subseteq \{\incp\} = \{\neg \conf\}$$

Hence, for any set of formulae $\Phi \subseteq \tlogic$:
    $$(\conf(\Phi) \leftrightarrow \neg \incp(\Phi)) \rightarrow \neg \incp(\Phi) \rightarrow \conp(\Phi) \rightarrow \neg \incf(\Phi)$$


From Definition \ref{def:tcvf}:\\
Let $\bM \in \mathtt{TMLN}$, an instantiation $I \subseteq \MI(\bM)$ and $x \in \{\conp, \conf, \incp, \incf\}$.

\hspace{-0cm}
\small{
\begin{tabular}{l}
    $\begin{array}{l} 
\Delta_{\conp}(I) =  
\left\{\begin{array}{ll}
1 & \text{ if $\conp(\TF(I))$ }  \\
0 & \text{ if $\neg \conp(\TF(I))$}
\end{array}\right.\end{array}$
 

$    \begin{array}{l} 
\Delta_{\conf}(I) =  
\left\{\begin{array}{ll}
1 & \text{ if $\conf(\TF(I))$ }  \\
0 & \text{ if $\neg \conf(\TF(I))$}
\end{array}\right.\end{array}$
\\

    $\begin{array}{l} 
\Delta_{\incp}(I) =  
\left\{\begin{array}{ll}
0 & \text{ if $\incp(\TF(I))$ }  \\
1 & \text{ if $\neg \incp(\TF(I))$}
\end{array}\right.\end{array}$
 
    $\begin{array}{l} 
\Delta_{\incf}(I) =  
\left\{\begin{array}{ll}
0 & \text{ if $\incf(\TF(I))$ }  \\
1 & \text{ if $\neg \incf(\TF(I))$}
\end{array}\right.\end{array}$
\end{tabular}
}

\normalsize
Therefore using the case equal to 1 and given that $(\conf(\Phi) \leftrightarrow \neg \incp(\Phi)) \rightarrow \neg \incp(\Phi) \rightarrow \conp(\Phi) \rightarrow \neg \incf(\Phi)$, for any instantiation $I \subseteq \MI(\bM)$:
$$\Delta_{\conf}(I) = \Delta_{\incp}(I) \leq \Delta_{\conp}(I) \leq \Delta_{\incf}(I)$$


\end{proof}

\vspace{0.5cm}

\begin{proof} [Theorem \ref{theo:ordrew}] 
Let  $x \in \{\conp, \conf, \incp, \incf\}$, $\bM \in \mathtt{TMLN}$ and $\forall I \subseteq \MI(\bM)$, we know from 
Proposition \ref{prop:ordre} that: 
$$\Delta_{\conf}(I) = \Delta_{\incp}(I) \leq \Delta_{\conp}(I) \leq \Delta_{\incf}(I)$$

From Definition \ref{def:TPS}, a temporal parametric semantics is defined as follows: 
$$\mathtt{TPS}(I) = \Delta(I) \cdot \Theta\Big(\bigsig(I)\Big). $$

Consequently, $\Delta(I)$ is a coefficient of the combination of $\bigsig$ and $\Theta$.
Thus, for any $I$, $\bigsig$ and $\Theta$, we may order the results according to the $\Delta(I)$ coefficient.

Finally, given that the MAP Inference returns the instantiation with the maximum strength, all strengths of the MAP inferences are equal.

Therefore, if we denote by: 
\begin{itemize}
    \item $\mathtt{TPS}_{\conf} = \tuple{\Delta_{\conf},\bigsig,\Theta}$, 
    $\mathtt{TPS}_{\incp} = \tuple{\Delta_{\incp},\bigsig,\Theta}$,
    \item $\mathtt{TPS}_{\conp} = \tuple{\Delta_{\conp},\bigsig,\Theta}$, 
    $\mathtt{TPS}_{\incf} = \tuple{\Delta_{\incf},\bigsig,\Theta}$.
\end{itemize}
Hence: 
$\forall I_{\conf} \in \mathtt{map}(\bM,\mathtt{TPS}_{\conf})$, $\forall I_{\incp} \in \mathtt{map}(\bM,\mathtt{TPS}_{\incp})$,
$\forall I_{\conp} \in \linebreak \mathtt{map}(\bM,\mathtt{TPS}_{\conp})$,
$\forall I_{\incf} \in \mathtt{map}(\bM,\mathtt{TPS}_{\incf})$, 

\vspace{0.1cm}
$~~~~~~~~~\mathtt{TPS}_{\conf}(I_{\conf}) = 
\mathtt{TPS}_{\incp}(I_{\incp}) \leq\mathtt{TPS}_{\conp}(I_{\conp}) \leq
\mathtt{TPS}_{\incf}(I_{\incf}).$

\end{proof}

\vspace{0.5cm}

\begin{proof} [Proposition \ref{prop:ThetaWB}] 
Let $\{w_1, \cdots, w_n\}$ such that $n \in [0, +\infty[$ and $\forall i \in [0,n]$, $w_i \in [0,1]$.
\begin{itemize}
    \item $\Theta_{sum}(w_1, \cdots, w_n) = \sum^{n}\limits_{i = 1} w_i$
    \begin{description} 
			\item [(a)] $\Theta_{sum}() = 0$: by definition.
			\item [(b)] $\Theta_{sum}(w_1) = w_1$: $\sum^{1}\limits_{i = 1} w_i = w_1$.
			\item [(c)] $\Theta_{sum}(w_1, \cdots, w_k) = \Theta_{sum}(w_1, \cdots, w_k, 0)$: $0$ is the identity element of the addition.  
			\item [(d)] $\Theta_{sum}(w_1, \cdots, w_k, y) \leq \Theta_{sum}(w_1, \cdots, w_k,z)$ 
			            if $y \leq z$: the addition is monotonic.
			\item [(e)] $\Theta_{sum}$ is symmetric: the addition is symmetric.
		\end{description}

    \item $\Theta_{sum,\alpha}(w_1, \cdots, w_n) = \left(\sum^{n}\limits_{i = 1} (w_i)^{\alpha}\right)^{\frac{1}{\alpha}}$ s.t. $\alpha \geq 1$
    \begin{description} 
			\item [(a)] $\Theta_{sum,\alpha}() = 0$: by definition.
			\item [(b)] $\Theta_{sum,\alpha}(w_1) = w_1$: $\left(\sum^{1}\limits_{i = 1} (w_i)^{\alpha}\right)^{\frac{1}{\alpha}} = w_i$.
			\item [(c)] $\Theta_{sum,\alpha}(w_1, \cdots, w_k) = \Theta_{sum,\alpha}(w_1, \cdots, w_k, 0)$: 0 cannot increase with any power and it is the identity element of the addition.  
			\item [(d)] $\Theta_{sum,\alpha}(w_1, \cdots, w_k, y) \leq \Theta_{sum,\alpha}(w_1, \cdots, w_k,z)$ 
			            if $y \leq z$: it is a sum of positive element and the addition is monotonic.
			\item [(e)] $\Theta_{sum,\alpha}$ is symmetric: the addition is monotonic.
		\end{description}
    
    \item $\Theta_{psum}(w_1, \cdots, w_n) = w_1 \ominus \cdots \ominus w_n$, \\
            where $w_1 \ominus w_2 = w_1 + w_2 - w_1 \cdot w_2$
            \begin{description} 
			\item [(a)] $\Theta_{psum}() = 0$: by definition.
			\item [(b)] $\Theta_{psum}(w) = w$: by definition.
			\item [(c)] $\Theta_{psum}(w_1, \cdots, w_k) = \Theta_{psum}(w_1, \cdots, w_k, 0)$: $w \ominus 0 = w + 0 - w \cdot 0 = w$.
			\item [(d)] $\Theta_{psum}(w_1, \cdots, w_k, y) \leq \Theta_{psum}(w_1, \cdots, w_k,z)$ 
			            if $y \leq z$: we know that any $w_i \in [0,1]$. From $w_1 \ominus w_2 = w_1 + w_2 - w_1 \cdot w_2$ then the addition of $w_1 + w_2$  is greater than the subtraction of the product $w_1 \cdot w_2$, i.e. $w_1 + w_2 \geq w_1 \cdot w_2$.
			\item [(e)] $\Theta_{psum}$ is symmetric: the addition and the product are symmetric.
		\end{description}
\end{itemize}
The three functions $\Theta_{sum}$, $\Theta_{sum, \alpha}$ and $\Theta_{psum}$ are well-behaved.
\end{proof}

\vspace{0.5cm}


\begin{proof} [Proposition \ref{prop:sigmaWB}] 
Let $\bM \in \mathtt{TMLN}$, $\{(\phi_1,w_1),\cdots,(\phi_n,w_n)\} \subseteq \MI(\bM)$.
\begin{itemize}
    \item $\bigsig_{id}(\{(\phi_1,w_1),\cdots,(\phi_n,w_n)\}) = (w_1, \cdots, w_n)$
    
    \begin{description} 
			\item [(a)] $\bigsig_{id}() = ()$: by definition. 
 			\item [(b)] $\bigsig_{id}(\{(\phi_1,w_1),\cdots,(\phi_k,w_k)\})$\\
 			$= (w'_1, \cdots,w'_n)$ such that if $k \geq 1$ then $n \geq 1$: $\forall k \geq 1$, $n = k$. 
 			\item [(c)] $\bigsig_{id}(\{(\phi_1,w_1),\cdots,(\phi_k,w_k),(\phi_{k+1},0)\})$ 
 			$= (\bigsig_{id}(\{(\phi_1,w_1),\cdots,(\phi_k,w_k)\}),0)$,\\
 			 if $\bigtau(\TF(\{\phi_1,\cdots,\phi_k\})) \not\vdash \bigtau(\{\phi\})$: 
 			 the function return the initial weight, then any weight doesn't change and the 0 is added.
 			 \item [(d)] $\bigsig_{id}(\{(\phi_1,w_1),\cdots,(\phi_k,w_k)\})$ 
 			$\subset  \bigsig_{id}(\{(\phi_1,w_1),\cdots,(\phi_k,w_k),(\phi_{k+1},w_{k+1})\})$ if
 			$\phi_{k+1}$ is a ground temporal formula, 
 			$\bigtau(\TF(\{\phi_1,\cdots,\phi_k\})) \not\vdash \bigtau(\{\phi_{k+1}\})$ and  
 			$\mathtt{Con}(\TF(\{(\phi_1,w_1),\cdots,(\phi_k,w_k),(\phi_{k+1},w_{k+1})\}))$: 
 			same reason as in item (c).
 			\item [(e)] for any well behaved $\Theta$:
 			$\Theta\Big(\bigsig_{id}(\{(\phi_1,w_1),\cdots,(\phi_k,w_k)\})\Big)$\\
 			$\leq \Theta\Big(\bigsig_{id}(\{(\phi_1,w_1),\cdots,(\phi_k,w_k),(\phi_{k+1},w_{k+1})\})\Big)$ if 
 			$\bigtau(\TF(\{\phi_1,\cdots,\phi_k\})) \not\vdash \bigtau(\{\phi_{k+1}\})$ 
 			and $\mathtt{Con}(\TF(\{(\phi_1,w_1),\cdots,(\phi_k,w_k),(\phi_{k+1},w_{k+1})\}))$: \\
 			same reason as in item (c) and (d).
		\end{description}

	\item $\bigsig_{thresh,\alpha}(\{(\phi_1,w_1),\cdots,(\phi_n,w_n)\}) = (\mathtt{max}(w_1 - \alpha, 0), \cdots, \mathtt{max}(w_n - \alpha, 0))$ s.t. $\alpha \in [0,1[$
	
	\begin{description} 
			\item [(a)] $\bigsig_{thresh,\alpha}() = ()$: by definition.
 			\item [(b)] $\bigsig_{thresh,\alpha}(\{(\phi_1,w_1),\cdots,(\phi_k,w_k)\})$ 
 			$= (w'_1, \cdots,w'_n)$ such that if $k \geq 1$ then $n \geq 1$: : $\forall k \geq 1$, $n = k$. 
 			\item [(c)] $\bigsig_{thresh,\alpha}(\{(\phi_1,w_1),\cdots,(\phi_k,w_k),(\phi_{k+1},0)\})$ 
 			$= \\ (\bigsig_{thresh,\alpha}(\{(\phi_1,w_1),\cdots,(\phi_k,w_k)\}),0)$, 
 			 if $\bigtau(\TF(\{\phi_1,\cdots,\phi_k\})) \not\vdash \bigtau(\{\phi\})$: \\
 			 every 0 still 0 and doesn't change the rest of the other weights.
 			 \item [(d)] $\bigsig_{thresh,\alpha}(\{(\phi_1,w_1),\cdots,(\phi_k,w_k)\})$ 
 			$\subset \bigsig_{thresh,\alpha}(\{(\phi_1,w_1),\cdots,(\phi_k,w_k),(\phi_{k+1},w_{k+1})\})$ if
 			$\phi_{k+1}$ is a grounded temporal formula, 
 			$\bigtau(\TF(\{\phi_1,\cdots,\phi_k\})) \not\vdash \bigtau(\{\phi_{k+1}\})$ and
 			$\mathtt{Con}(\TF(\{(\phi_1,w_1),\cdots,(\phi_k,w_k),(\phi_{k+1},w_{k+1})\}))$.\\
 			Any weight is added.
 			\item [(e)] for any well behaved $\Theta$: 
 			$\Theta\Big(\bigsig_{thresh,\alpha}(\{(\phi_1,w_1),\cdots,(\phi_k,w_k)\})\Big)$\\
 			$\leq \Theta\Big(\bigsig_{thresh,\alpha}(\{(\phi_1,w_1),\cdots,(\phi_k,w_k),(\phi_{k+1},w_{k+1})\})\Big)$ if 
 			$\bigtau(\TF(\{\phi_1,\cdots,\phi_k\})) \not\vdash \bigtau(\{\phi_{k+1}\})$ 
 			and $\mathtt{Con}(\TF(\{(\phi_1,w_1),\cdots,(\phi_k,w_k),(\phi_{k+1},w_{k+1})\}))$.\\
            Any weight is added and the previous weights are independent to new weight then the aggregation cannot decrease. 			
		\end{description}

    \item 
    \begin{itemize}
        \item let $\phi = (\psi_1 \wedge \cdots \wedge \psi_k) \rightarrow \psi_{k+1}$ a rule, \\ $\mathtt{prem}(\phi) = \{\psi_1, \cdots, \psi_k\}$.
        
        \item $\mathtt{imp}((\phi,w),\{(\phi_1,w_1), \cdots, (\phi_n,w_n)\}) = $ 
        $\begin{array}{l} 
                  \left\{\begin{array}{ll}
                 0 & \text{ if $\phi$ is a ground rule s.t.
                 $\exists \psi_i \in \mathtt{prem}(\phi)$ }\\
                  & \text{s.t. $\psi_i \notin \CN(\{\phi_1, \cdots,\phi_n\})$ }  \\
                 w & \text{ otherwise}
                 \end{array}\right.\end{array}$
                 
        \item $\bigsig_{rule}(\{(\phi_1,w_1),\cdots,(\phi_n,w_n)\}) = \Big(\mathtt{imp}((\phi_1,w_1),\{(\phi_2,w_2), \cdots, (\phi_n,w_n)\}), \cdots,$\\ $\mathtt{imp}((\phi_n,w_n),\{(\phi_1,w_1), \cdots, (\phi_{n-1},w_{n-1})\}) \Big) $ 
    \end{itemize}
    
    \begin{description} 
			\item [(a)] $\bigsig_{rule}() = ()$: by definition.
 			\item [(b)] $\bigsig_{rule}(\{(\phi_1,w_1),\cdots,(\phi_k,w_k)\})$ 
 			$= (w'_1, \cdots,w'_n)$ such that if $k \geq 1$ then $n \geq 1$.  
 			Any new weight is either added with its weight or with 0 but it is added then $\forall k \geq 1$, $n = k$.
 			\item [(c)] $\bigsig_{rule}(\{(\phi_1,w_1),\cdots,(\phi_k,w_k),(\phi_{k+1},0)\})$ 
 			$= (\bigsig_{rule}(\{(\phi_1,w_1),\cdots,(\phi_k,w_k)\}),0)$, 
 			 if $\bigtau(\TF(\{\phi_1,\cdots,\phi_k\})) \not\vdash \bigtau(\{\phi\})$.\\
 			 Add a weighted temporal formula with a weight of 0 will still be 0 in any case.
 			 \item [(d)] $\bigsig_{rule}(\{(\phi_1,w_1),\cdots,(\phi_k,w_k)\})$ 
 			$\subset \bigsig_{rule}(\{(\phi_1,w_1),\cdots,(\phi_k,w_k),(\phi_{k+1},w_{k+1})\})$ if
 			$\phi_{k+1}$ is a ground temporal formula, 
 			$\bigtau(\TF(\{\phi_1,\cdots,\phi_k\})) \not\vdash \bigtau(\{\phi_{k+1}\})$ and
 			$\mathtt{Con}(\TF(\{(\phi_1,w_1),\cdots,(\phi_k,w_k),(\phi_{k+1},w_{k+1})\}))$.\\
 			Add any new consistent information doesn't change the other information then we have a strict inclusion.
 			\item [(e)] for any well behaved $\Theta$: 
 			$\Theta\Big(\bigsig_{rule}(\{(\phi_1,w_1),\cdots,(\phi_k,w_k)\})\Big)$\\
 			$\leq \Theta\Big(\bigsig_{rule}(\{(\phi_1,w_1),\cdots,(\phi_k,w_k),(\phi_{k+1},w_{k+1})\})\Big)$ if 
 			$\bigtau(\TF(\{\phi_1,\cdots,\phi_k\})) \not\vdash \bigtau(\{\phi_{k+1}\})$ 
 			and $\mathtt{Con}(\TF(\{(\phi_1,w_1),\cdots,(\phi_k,w_k),(\phi_{k+1},w_{k+1})\}))$.\\
 			Add a new information has a weight belonging to 0 and 1, it cannot decrease the aggregation.
		\end{description}

\end{itemize}
The three functions $\bigsig_{id}$, $\bigsig_{thresh,\alpha}$ and $\bigsig_{rule}$ are well-behaved.
\end{proof}

\end{appendix}


\end{document}